# Bangla Word Clustering Based on Tri-gram, 4-gram and 5-gram Language Model

**Dipaloke Saha**[1,*], **Md Saddam Hossain**[1], **MD. Saiful Islam**[1] **and Sabir Ismail**[1]

[1]Department Of Computer Science & Engineering,
Shahjalal University Of Science And Technology, Sylhet, Bangladesh.
dipsustcse12@gmail.com, mshossaincse@gmail.com , saiful-cse@sust.edu ,
sabir.ismail01@gmail.com

*Keywords:*

- *Word Cluster, Natural Language Processing, Machine Learning, N-gram Model, Term Frequency (tf).*
- *SUST, ICERIE.*

**Abstract**: — In this paper, we describe a research method that generates Bangla word clusters on the basis of relating to meaning in language and contextual similarity. The importance of word clustering is in parts of speech (POS) tagging, word sense disambiguation, text classification, recommender system, spell checker, grammar checker, knowledge discover and for many others Natural Language Processing (NLP) applications. In the history of word clustering, English and some other languages have already implemented some methods on word clustering efficiently. But due to lack of the resources, word clustering in Bangla has not been still implemented efficiently. Presently, it's implementation is in the beginning stage. In some research of word clustering in English based on preceding and next five words of a key word they found an efficient result. Now, we are trying to implement the tri-gram, 4-gram and 5-gram model of word clustering for Bangla to observe which one is the best among them. We have started our research with quite a large corpus of approximate 1 lakh Bangla words. We are using a machine learning technique in this research. We will generate word clusters and analyze the clusters by testing some different threshold values.

## 1. INTRODUCTION

Though Bangla is a widely spoken language, it has lack of resources in its research field. Recently, a new research dimension in Bangla is added called word clustering. In this paper, the research of word clustering for Bangla language is trying to be extended. For this, a large Bangla corpus containing 97,971 individual words is compiled to generate the word clusters. In this paper, an unsupervised machine learning technique and a method are proposed to cluster Bangla words on the basis of similarity in semantics and contexts.

In language processing word cluster has a wide range of applications. POS tag is one of them. Same clustered words usually contain the same POS tag. Word clustering can produce suggestions for an inaccurately typed word which is very much helpful for spell checker. Word sense disambiguation, sentence structure with grammatical mistakes can also be solvable using clustered words. In the case of recommender system if related products of the same category are clustered in the same group, more feasible suggestion can be produced. This type of work is also useful for Bangla search engine to find the appropriate content. So, there is a huge importance of word clustering in the field of natural language processing.

[*] dipsustcse12@gmail.com

## 2. RELATED WORK

In Bangla the implementation of word clustering is in the neophyte stage. A previous work on Bangla word clustering exists in which an unsupervised machine learning technique is used to implement the bigram model by Sabir Ismail and M. Shahidur Rahman. In many other languages different types of techniques are used for word clustering. Finch and Chater (1992) implemented bigram model for the calculation of weight matrix of a neural network. N-gram language model is used on word clustering in a research proposed by Brown, Desouza, Mercer, Peitra, Lai (1992). Another effort using n-gram model is introduced by Korkmaz (1997) in which a similarity function and greedy algorithm is used to group the words into same cluster. However, with the use of delete interpolation method by Mori, Nishimura and Itoh (1998) they got the better result than the Brown, Desouza's method. This was done for Japanese and English language. Besides these, there exists quite a good number of researches of word clustering for some other languages like Russian, Arabic, Chinese etc.

## 3. PROBLEM DEFINITION

Clustering is an unsupervised machine learning technique that does not require any type of rules or predefined conditions. Items which are much similar either in semantically or contextually are grouped in the same cluster and which are dissimilar are in different clusters. The introduced method in this problem is concentrating on two types of similarity such as semantics and contextual similarity.

Consider the following four sentences:

1. মা বাবার দেখাশুনা করা প্রত্যেক সন্তানের দায়িত্ব।
2. মা বাবার দেখভাল করা প্রত্যেক সন্তানের দায়িত্ব।
3. গতরাতে ভালুকায় কয়েকজন লোক নিহত হয়।
4. গতরাতে ভালুকায় কয়েকজন লোক আহত হয়।

দেখাশুনা and দেখভাল are similar in semantic meaning in sentence 1 and 2 and there is similarity in নিহত and আহত in sentence 3 and 4. Here, the theory of N-gram model is implemented. Probability distribution is used here to define n-th item in a sequence form previous or next (n-1) items. Tri-gram, $4^{th}$ and $5^{th}$ gram model is defined as size of 3, 4 and 5 of N-gram respectively. In this research, word clusters will be generated by implementing tri, $4^{th}$ and $5^{th}$ gram model. After finding the word clusters the most efficient model will be found out based on those clustering words.

## 4. METHODOLOGY

Firstly, quite a large corpus of 97,971 individual words $W_i$ is used in this research. Next, a list of previous three words of a specific word for tri-gram, four words for 4-gram, five words for 5-gram are prepared. Similarly, a list of next three words of a specific word for tri-gram, four words for 4-gram, five words for 5-gram are prepared. Next, similarity between a pair of words to be included in the same cluster based on preceding three words, four words and five words are determined as follows:

In tri-gram for every pair of words $W_i$, $W_j$ the number of matched preceding words from list
*list(Wi-3, Wi-2 , Wi-1) and list(Wj-3, Wj-2, Wj-1)*

*P(Wi,Wj)=(Count(match(list(Wi-3,Wi-2,Wi-1),list(Wj-3,Wj-2,Wj-1)))/((Count(list(Wi-3,Wi-2,Wi-1))+Count(list(Wj-3,Wj-2,Wj-1 )))*

Similarly, calculation for the 4-gram model is:

*P(Wi,Wj) = (Count(match(list(Wi-4,Wi-3,Wi-2 ,Wi-1),list(Wj-4,Wj-3,Wj-2,Wj-1)))/((Count(list(Wi-4,Wi-3,Wi-2,Wi-1))+Count(list(Wj-4,Wj-3,Wj-2,Wj-1)))*

and for 5-gram model is:

$P(W_i, W_j) = (Count(match(list(W_{i-5}, W_{i-4}, W_{i-3}, W_{i-2}, W_{i-1}), list(W_{j-5}, W_{j-4}, W_{j-3}, W_{j-2}, W_{j-1})))/((Count(list(W_{i-5}, W_{i-4}, W_{i-3}, W_{i-2}, W_{i-1})) + Count(list(W_{j-5}, W_{j-4}, W_{j-3}, W_{j-2}, W_{j-1})))$

Again similarly, between a pair of words to be included in the same cluster based on following three, four and five words are determined as follows,
For tri-gram,

$P(W_i, W_j) = (Count(match(list(W_{i+3}, W_{i+2}, W_{i+1}), list(W_{j+3}, W_{j+2}, W_{j+1})))/((Count(list(W_{i+3}, W_{i+2}, W_{i+1})) + Count(list(W_{j+3}, W_{j+2}, W_{j+1})))$

Similarly, calculation for the 4-gram model is:

$P(W_i, W_j) = (Count(match(list(W_{i+4}, W_{i+3}, W_{i+2}, W_{i+1}), list(W_{j+4}, W_{j+3}, W_{j+2}, W_{j+1})))/((Count(list(W_{i+4}, W_{i+3}, W_{i+2}, W_{i+1})) + Count(list(W_{j+4}, W_{j+3}, W_{j+2}, W_{j+1})))$

and for 5-gram model is:

$P(W_i, W_j) = (Count(match(list(W_{i+5}, W_{i+4}, W_{i+3}, W_{i+2}, W_{i+1}), list(W_{j+5}, W_{j+4}, W_{j+3}, W_{j+2}, W_{j+1})))/((Count(list(W_{i+5}, W_{i+4}, W_{i+3}, W_{i+2}, W_{i+1})) + Count(list(W_{j+5}, W_{j+4}, W_{j+3}, W_{j+2}, W_{j+1})))$

If the above equations of a particular model yield values greater than a predefined threshold value they are grouped into the same cluster for that model.
For example, to implement the tri-gram model some of the following phrases are :

1. ভোরে সূর্য উঠার আগে।
2. আগে খাওয়া শেষ করি।
3. সকালে সূর্য উঠার পরে।
4. পরে কাজটি শেষ করি।

For word আগে preceding three words list:

$list(W_{i-3}, W_{i-2}, W_{i-1})$ = { ভোরে , সূর্য , উঠার }
 $Count(list(W_{i-3}, W_{i-2}, W_{i-1})) = 3$

For word আগে Following three words list :
 $list(W_{i+3}, W_{i+2}, W_{i+1})$ = { খাওয়া, শেষ , করি }
 $Count(list(W_{i+3}, W_{i+2}, W_{i+1})) = 3$

For word পরে preceding three words list:
 $list(W_{j-3}, W_{j-2}, W_{j-1})$ = { সকালে , সূর্য , উঠার }
 $Count(list(W_{j-3}, W_{j-2}, W_{j-1})) = 3$

For word পরে following three words list:

 $list(W_{j+3}, W_{j+2}, W_{j+1})$ = { কাজটি , শেষ , করি }
 $Count(list(W_{j+3}, W_{j+2}, W_{j+1})) = 3$

Number of matched words for word আগে with পরে based on preceding three words :

 $Count(match(list(W_{i-3}, W_{i-2}, W_{i-1}), list(W_{j-3}, W_{j-2}, W_{j-1}))) = 2$
 $Count(list(W_{i-3}, W_{i-2}, W_{i-1})) + Count(list(W_{j-3}, W_{j-2}, W_{j-1})) = 6$

Similarity between words আগে and পরে based on preceding three words:
   *P(Wi, Wj) = 2/6 = 0.33*

Number of matched words for word আগে and পরে based on following three words:

   *Count(match(list(Wi+3, Wi+2 , Wi+1),list(Wj+3, Wj+2 , Wj+1))) = 2*
   *Count(list(Wi+3, Wi+2, Wi+1)) + Count(list(Wj+3, Wj+2, Wj+1)) = 6*

   Similarity between words আগে and পরে based on following three words:
   P(Wi, Wj) = 2/6 = 0.33

Similarly, 4$^{th}$ and 5$^{th}$ gram model can be implemented in the same way.

The value of similarity between words আগে with পরে when considering preceding three words is 0.33 and considering following three words it is also 0.33. Different types of threshold values are experimented and best result is earned with 0.20. Both words are grouped in the same cluster when all the probability scores are greater than this threshold value.

## 5. RESULT ANALYSIS

In the tri, 4$^{th}$ and 5$^{th}$ gram model we derive 2215, 3327 and 5730 word clusters in total respectively. Some clusters randomly from each of the model are represented here in the following tables:

Table 1  Word Cluster for tri-gram model

| | |
|---|---|
| ফায়ার সার্ভিস | মেডিকেল কলেজ |
| মুখোমুখি সংঘর্ষ | দেখাশোনা দেখভাল |
| উপপরিদর্শক এসআই | সহকারিসহ সঙ্গী |
| পিকআপকে রিকশাকে | প্রথম আলোকে |
| বর্ডার গার্ড | মহানগর প্রভাতী |
| প্রেমের সম্পর্ক | আশ্বাসের পূরণের |
| ফেরদৌস জুনায়েদ | বিচারিক হাকিম |

Table 2  Word Cluster for 4th gram model

| | |
|---|---|
| লাখ<br>টাকা | নায়েক<br>সুবেদার |
| সহকারিসহ<br>সঙ্গী | চালবোঝাই<br>বালুবোঝাই |
| মহানগর<br>প্রভাতী | দুমড়ে<br>মুচড়ে |
| কমিটি<br>গঠন | মালামালসহ<br>আসবাবপত্রসহ |
| ঘটনাস্থল<br>পরিদর্শন | দেশলাইয়ের<br>কাঠি |
| বার্ষিক<br>ওয়াজ | ফায়ার<br>সার্ভিস |

Table 3  Word Cluster for 5th gram Model

| | |
|---|---|
| ওরস<br>মাহফিল | স্টুডেন্ট<br>ভিসায় |
| মিনার<br>মসজিদ | আশঙ্কায়<br>বিঘ্ন |
| ঘটনাস্থল<br>পরিদর্শন | আশ্বাসের<br>পূরণের |
| গোলা<br>অবিস্ফোরিত | মার্কেটিং<br>ম্যানেজার |
| বিপর্যয়<br>চরম | মেরে<br>পিটিয়ে |
| এনটিভি<br>আরটিভি | গার্লস<br>ক্যাডেট |

After analyzing the word clusters of all the three models we find poor similarity in some word clusters such as 266 for tri-gram, 300 for 4$^{th}$ gram and 360 for 5$^{th}$ gram. So, we find 1949, 3027 and 5370 clusters in strong similarity for the tri, 4$^{th}$ and 5$^{th}$ gram model respectively. So, the accuracy for strong similarity in

Tri-gram :- 88%

4$^{th}$gram :- 91%

5$^{th}$gram :- 93%

So, it is observed that 4$^{th}$ gram is is better than tri-gram and 5$^{th}$ gram is the best in all of them.

## 6. CONCLUSION

Word clustering is important for various types of purpose for any language. For this reason in Bangla, tri-gram, 4$^{th}$ gram and 5$^{th}$ gram model is implemented here to proceed the previous work on word clustering. The analysis and result presented above on quite a large Bangla corpus has helped us to find the efficiency among the three mentioned models for word clustering. On the basis of the observation, it can be said that better efficiency is in the higher orders than the preceding orders of the N-gram model.